\documentclass{article}
\usepackage[T1]{fontenc}
\usepackage[preprint]{colm2025_conference} 
\usepackage{xcolor}
\usepackage{hyperref}
\hypersetup{hidelinks}
\usepackage[margin=1.25in]{geometry}
\usepackage{graphicx}
\usepackage{amsmath}
\usepackage{amsfonts}
\usepackage{wrapfig}
\usepackage{amssymb}
\usepackage{adjustbox}
\usepackage{enumitem}
\usepackage{booktabs}
\usepackage{wrapfig}
\usepackage{needspace}
\usepackage{amsthm}
\usepackage[ruled,linesnumbered,noresetcount]{algorithm2e}
\usepackage{mathtools}
\usepackage{csquotes}
\usepackage{array}
\usepackage{url,eucal}
\usepackage{tabularx}
\usepackage{makecell}
\usepackage{xspace}
\usepackage{multirow}

\theoremstyle{definition}

\newcommand{\InternW}{InternW\xspace}
\newcommand{\InternWZero}{InternW0\xspace}

\newcommand{\EgoLab}{EgoLab\xspace}

\title{\InternWZero: A Foundational Physical World Model for Efficient Real-World Interactions}
\author{Physical Intelligence Team, Shanghai AI Laboratory}
\date{September 2026}

\begin{document}
\maketitle

\begin{figure}[ht]
    \centering
    \vspace{-15pt}
    \includegraphics[width=0.96\textwidth]{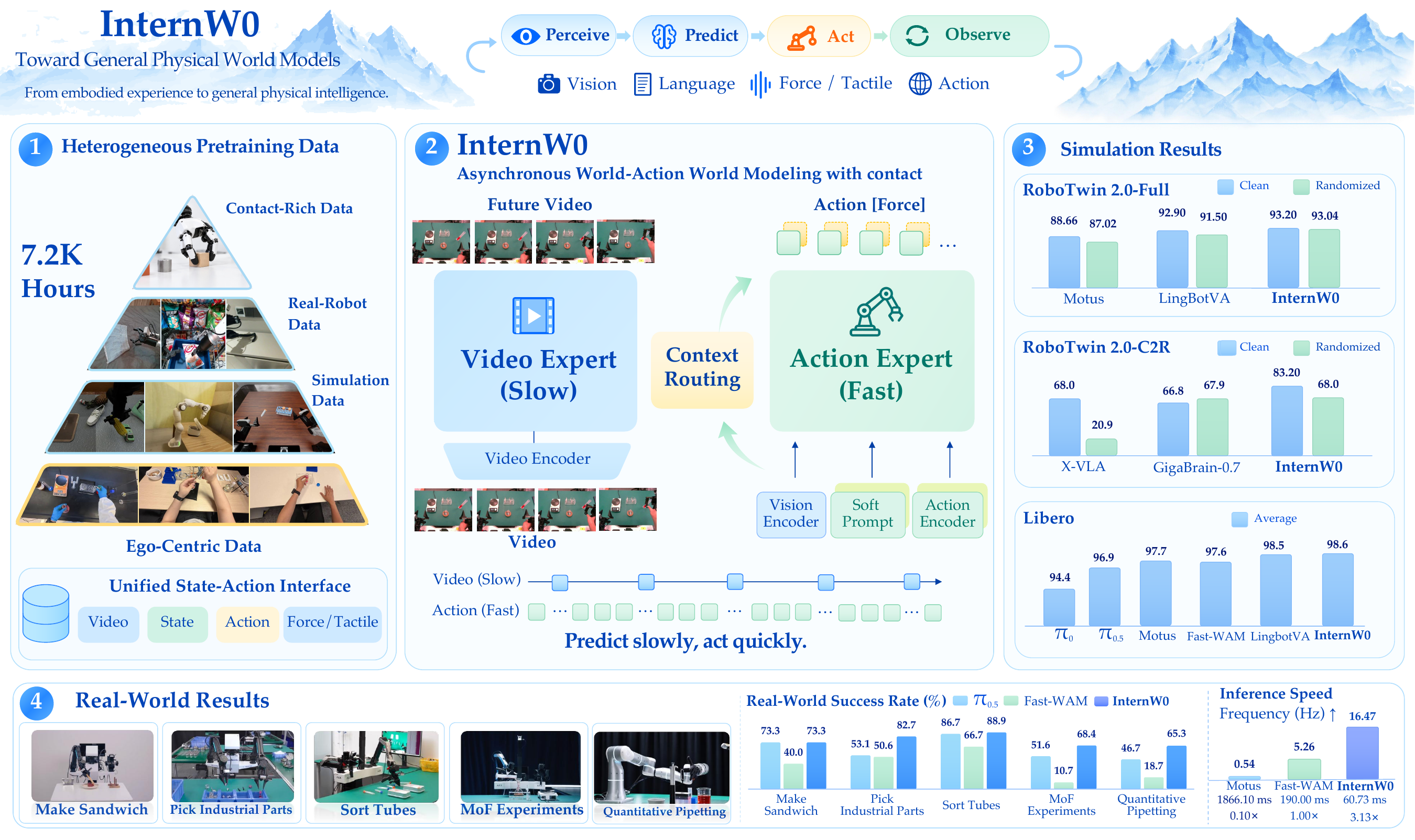}
    \label{fig:w-teaser}
\end{figure}

\begin{abstract}
Physical intelligence requires more than predicting how the world may
evolve: predictions must remain actionable as the world continues to
change. We introduce \InternWZero, the first instantiation of the
\InternW physical world model series from Shanghai AI Laboratory,
built around omnimodal interfaces, asynchronous multi-frequency
processing, and local physical modeling under partial observations and
external influences. \InternWZero jointly learns future visual dynamics and continuous robot
control through an asymmetric video--action architecture with flow
matching. A high-capacity video expert provides longer-horizon
predictive context, while a lightweight action expert operates at a
faster timescale. Instead of regenerating the future for every action
update, \InternWZero reuses layerwise K/V and adapts it
to newly observed states through observation-conditioned context routing.
Domain-specific interfaces and soft prompts support heterogeneous
embodiments, while contact-aware post-training incorporates force and
tactile signals for contact-rich manipulation.

We train \InternWZero on approximately 7,200 hours of heterogeneous
robot and egocentric data, including EgoLab, a 275-hour
real-laboratory egocentric dataset. Evaluation spans simulation benchmarks and real-world scientific tasks, including a
15-stage metal--organic framework synthesis workflow and 5-stage contact- and force-aware dexterous manipulation for general-purpose quantitative pipetting. These results advance scalable, asynchronous,
and science-native physical world models for universal and efficient real-world interactions.

\vspace{10pt}

\textbf{Project page:} \url{https://internrobotics.github.io/internw0}
\end{abstract}

\section{Design Motivation and Overall Framework}\label{sec:introduction}


Physical intelligence concerns an AI agent's ability to interact with the physical world, requiring it to connect its understanding of the environment and the task with the consequences of its actions. This connection demands more than recognizing objects or interpreting instructions: the agent must estimate the current physical state, anticipate how candidate actions will change it, and use feedback to revise its decisions. These capabilities must operate under partial observations~\citep{kaelbling1998pomdp} and limited computation, even as the environment continues to evolve. A world model establishes this connection by representing how physical states evolve in response to actions, enabling the agent to translate its understanding of the environment and the task into informed decisions~\citep{ha2018worldmodels,hafner2025dreamerv3}.

Recent advances in physical world modeling have increasingly focused on
world action models (WAMs), which couple future
visual prediction with robot action generation and provide a promising
way to ground predictive dynamics in executable interaction~\citep{wang2026world, yuan2026fastwam, bi2026motus, ye2026dreamzero}. A unified WAM is expected to support heterogeneous embodiments and richer physical modalities such as proprioception, force, and tactile feedback, while avoiding the latency introduced by synchronously updating expensive video prediction and fast action generation. However, these
remain insufficiently addressed by existing WAMs.

In this report, 
we develop the \InternW world model series to provide this connection through a shared model of perception, physical dynamics, and action. Following the resource-constrained perspective from Shanghai AI Laboratory \citep{chen2026definition}, a physical world model is viewed as a compact approximation of physical state transitions under finite sensing and computational resources. This perspective motivates three requirements of world modeling: omnimodal perception and interaction, asynchronous processing at multiple frequencies, and local environment modeling under external influences.

\begin{itemize}[leftmargin=2.2em]
    \item \textbf{Omnimodal interfaces:} \InternW aims to unify diverse modality inputs and predict actions within a shared physical representation, enabling embodied agents to perceive, reason, and interact with a wide range of physical environments.
    \item \textbf{Asynchronous and multi-frequency signal processing:} \InternW accommodates different sensing and control frequencies to balance prediction quality, responsiveness, and computational cost.
    \item \textbf{Local environment and action modeling:} \InternW estimates local physical state from partial observations and predicts the effects of both agent actions and external disturbances, updating its estimates as new evidence arrives.
\end{itemize}
    
\paragraph{Shared Representation and Specialized Experts.}
These requirements motivate the \InternW architecture described here at the series level. We draw on the modality-specific parameterization principle of mixture-of-transformers (MoT)~\citep{liang2025mixtureoftransformers}. Our series-level design adapts this principle to modalities (or data channels) for language semantics, visual dynamics, spatial geometry, contact mechanics, and body actions. Modality-specific processing thus supports joint reasoning about task intent, object configuration, contact conditions, and executable motion. The cross-modality interfaces are extensible, allowing individual \InternW models to select the modalities (or data channels) and computational capacity appropriate for their deployment setting.

\paragraph{Latent Dynamics for Prediction and Action.}
At the center of the framework, latent dynamics connect the estimated physical state to possible future states and interactions. Learning latent dynamics for planning from visual observations has been explored in prior model-based control work~\citep{hafner2019planet}. To express this role conceptually, let $\boldsymbol{\mathcal{H}}_t$ contain the timestamped observations and executed actions available up to time $t$, and let $\boldsymbol{z}_t$ denote a compact representation of local state and its uncertainty. We write
\begin{equation}
    \boldsymbol{z}_t = F_{\theta}(\boldsymbol{\mathcal{H}}_t),
    \qquad
    p_{\theta}(\boldsymbol{z}_{t+\Delta} \mid \boldsymbol{z}_t, \boldsymbol{u}_{t:t+\Delta}, \Delta),
    \label{eq:internw-local-dynamics}
\end{equation}
where $\boldsymbol{u}_{t:t+\Delta}$ denotes a candidate action sequence and $\Delta$ is the prediction horizon. Unobserved external influences contribute to the uncertainty of the transition distribution. This formulation specifies the modeling role rather than prescribing a particular probabilistic implementation. Prediction interfaces translate the latent representation into future observations or task-relevant states, while action interfaces generate commands conditioned on the task and available physical information. Their shared representation connects what the agent expects to happen with what it chooses to do. Under finite computational resources, we seek latent predictions that preserve physical information relevant to task outcomes and action choices, consistent with control-centric world modeling~\citep{hansen2024tdmpc2}. In the \InternW framework, we additionally require predictive information to become available in time to influence ongoing execution. We therefore regard predictive accuracy, decision relevance, and timeliness as three design criteria for our framework.

\paragraph{Duplex Interaction and Closed-Loop Feedback.}
\InternW couples perception, prediction, and action through duplex interaction: it continues to receive new observations while generating predictions and taking actions. Duplex interaction describes this concurrent input--output flow, whereas asynchronous, multi-frequency processing determines when each sensor stream and computational component is updated. Together, these mechanisms allow visual, contact, and body-state feedback to refine subsequent outputs while an operation is in progress. Each executed action produces new observations that update the local physical representation and inform the next prediction--action cycle. In scientific workflows, this closed loop connects experimental operations to their measured outcomes and supplies evidence for subsequent model refinement. Correct operation requires timestamp alignment and causal consistency: each update must account for actions that have already been executed while remaining able to revise commands that have not yet been issued.

\begin{wrapfigure}{r}{0.55\textwidth}
\centering
\includegraphics[width=0.5\textwidth]{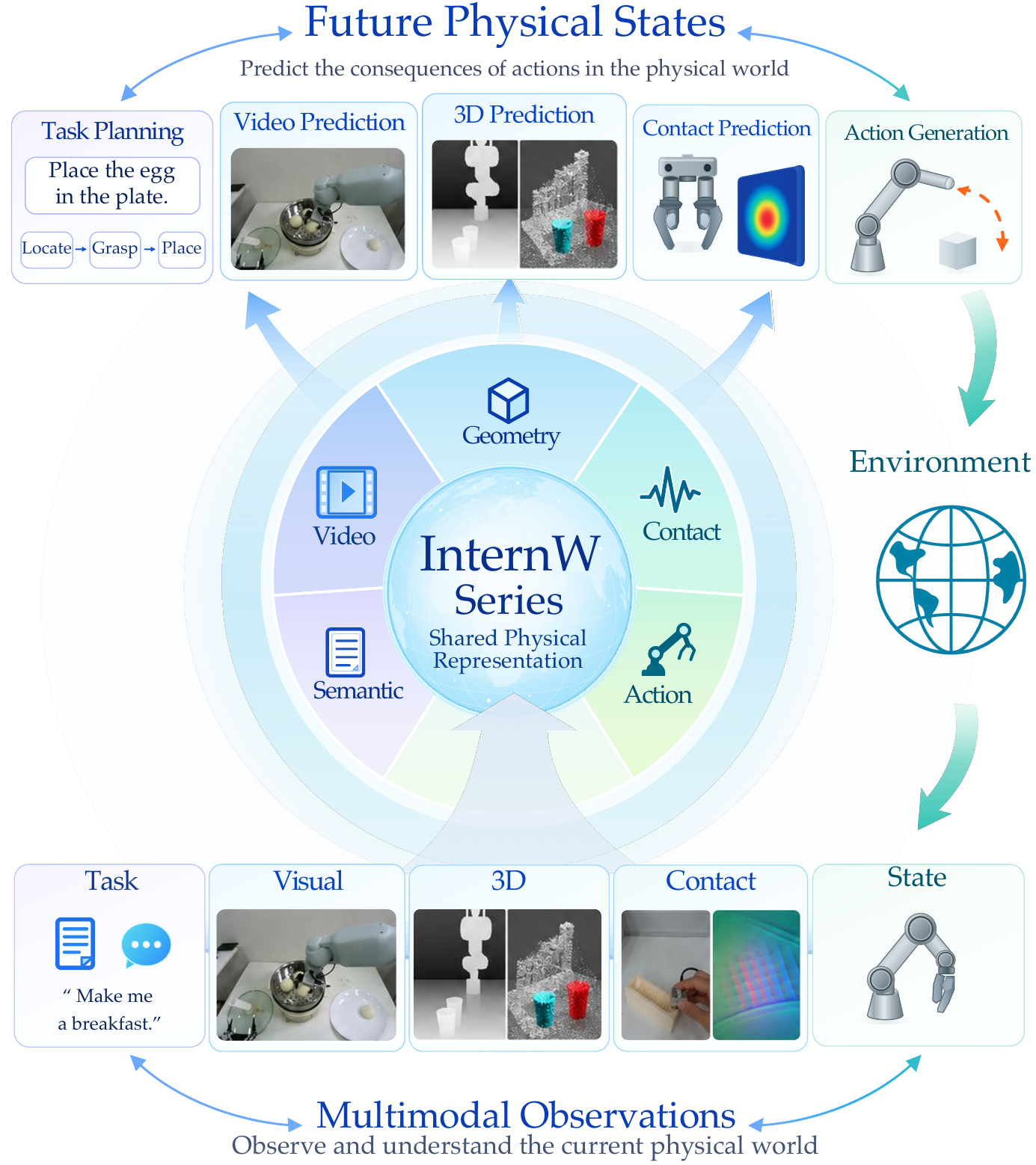}
\caption{\textbf{Framework of \InternW world model series.}}
\label{fig:w-frame}
\end{wrapfigure}

\paragraph{Multimodal Feedback for Contact-Aware Manipulation.}
\InternW is designed to incorporate force, tactile, and proprioceptive feedback into action generation alongside visual predictive context. These physical modalities provide complementary information about contact and the robot's physical state, including changes that may not be directly observable in images~\citep{lee2019visiontouch,chen2023vtt,song2026visual,yu2026forcevla}. Because these signals arrive at different temporal resolutions, the model separates long-horizon predictive context from recent feedback used to update each action. This enables local action correction at every feedback step without requiring the full predictive representation to be recomputed. Specifically, \InternWZero fuses the temporal history of interaction forces with visual and proprioceptive observations, while jointly predicting the end-effector poses and six-dimensional interaction wrenches. Each wrench comprises three force and three moment components~\citep{lynch2017modern}. The historical force inputs encode observed contact, whereas the predicted wrenches represent anticipated interaction loads. Together, they provide information about both past contact and anticipated loads to support contact-aware action selection. Active hybrid position--force interaction is a core manipulation capability that coordinates motion with contact-force regulation~\citep{raibert1981hybrid,li2026forcevla2}.

\paragraph{\InternWZero as an Initial Instantiation.}
Building on the above design principles, \InternWZero connects physical prediction and responsive interaction through a mixture-of-transformers (MoT) architecture. Specialized video and action experts are jointly optimized through video--action flow matching, coupling future visual dynamics with executable motion while preserving distinct processing pathways. Within the action expert, force feedback is integrated with visual and proprioceptive observations, enabling changes in contact and body state to condition subsequent action updates. This fusion provides a direct pathway from real-time sensory feedback to local motion adjustment. To coordinate future prediction with real-time observation, an observation-conditioned context-routing interface uses current visual features to query predictive video representations and construct chunk-specific context for the action expert. The resulting duplex interaction allows longer-horizon predictive context to guide execution while incoming observations continually inform subsequent actions, without requiring a new prediction cycle for every interaction update. Domain-specific state and action encoders, action decoders, and soft prompts adapt this shared architecture to heterogeneous datasets and embodiments. These mechanisms concretely realize the framework's multimodal interfaces, asynchronous processing, and observation-driven local updates. Within the InkStone scientific discovery platform~\citep{InkStone}, our system connects the scientific reasoning and tool-use capabilities of Intern-S2-Preview~\citep{bai2026interns2preview} to physical experimentation through \InternWZero. Execution records and measured outcomes can support subsequent evaluation and model improvement. The following sections detail the architecture and training pipeline, data recipe, simulation benchmarks, real-robot and wet-lab experiments, as well as the future work of InternW world model series.

\section{\InternWZero Architecture and Training Pipeline}\label{sec:definitions}

\subsection{Architecture}
\label{sec:w0-architecture}

\InternWZero instantiates the preceding framework with a
\emph{mixture-of-transformers} (MoT) backbone comprising a video expert
for future visual prediction and a lightweight action expert for robot
control. Each joint layer contains modality-specific blocks with
separate parameters and token streams, connected through an
observation-conditioned video-context interface. This structure allows
the two experts to use different capacities while coupling visual
prediction with action generation.

The central design goal is to keep future predictions useful as new
sensory signals arrive during execution. This requires coordinating
prediction and control at different timescales: repeatedly recomputing
a complete predictive plan at every control update is expensive,
whereas executing against an unchanged plan context risks conditioning the
policy on an increasingly stale view of the world.

\InternWZero addresses this tension through \emph{asynchronous duplex
inference}, in which predictive video modeling and action generation
operate at different temporal scales while remaining coupled through
an observation-conditioned context interface. The heterogeneous
pretraining stage learns this coupling from visual observations,
proprioceptive states, and robot actions. For contact-rich downstream
tasks, the action interface is further extended during post-training
with force and tactile observations and joint prediction of future
interaction signals. Embodiment-specific interfaces and soft prompts
support transfer across heterogeneous control spaces and sensing
configurations.

The interaction is intentionally asymmetric. The video expert builds a
shared predictive representation from visual observations and language
without consuming robot-specific action tokens or domain identities.
The action expert, in turn, reads the video context 
together with the latest observation, proprioception, language
conditioning, and embodiment-specific interfaces. This separation keeps
physical prediction broadly shared while allowing control to specialize
across heterogeneous embodiments.

\begin{figure}[ht]
  \centering
  \includegraphics[width=0.95\textwidth]{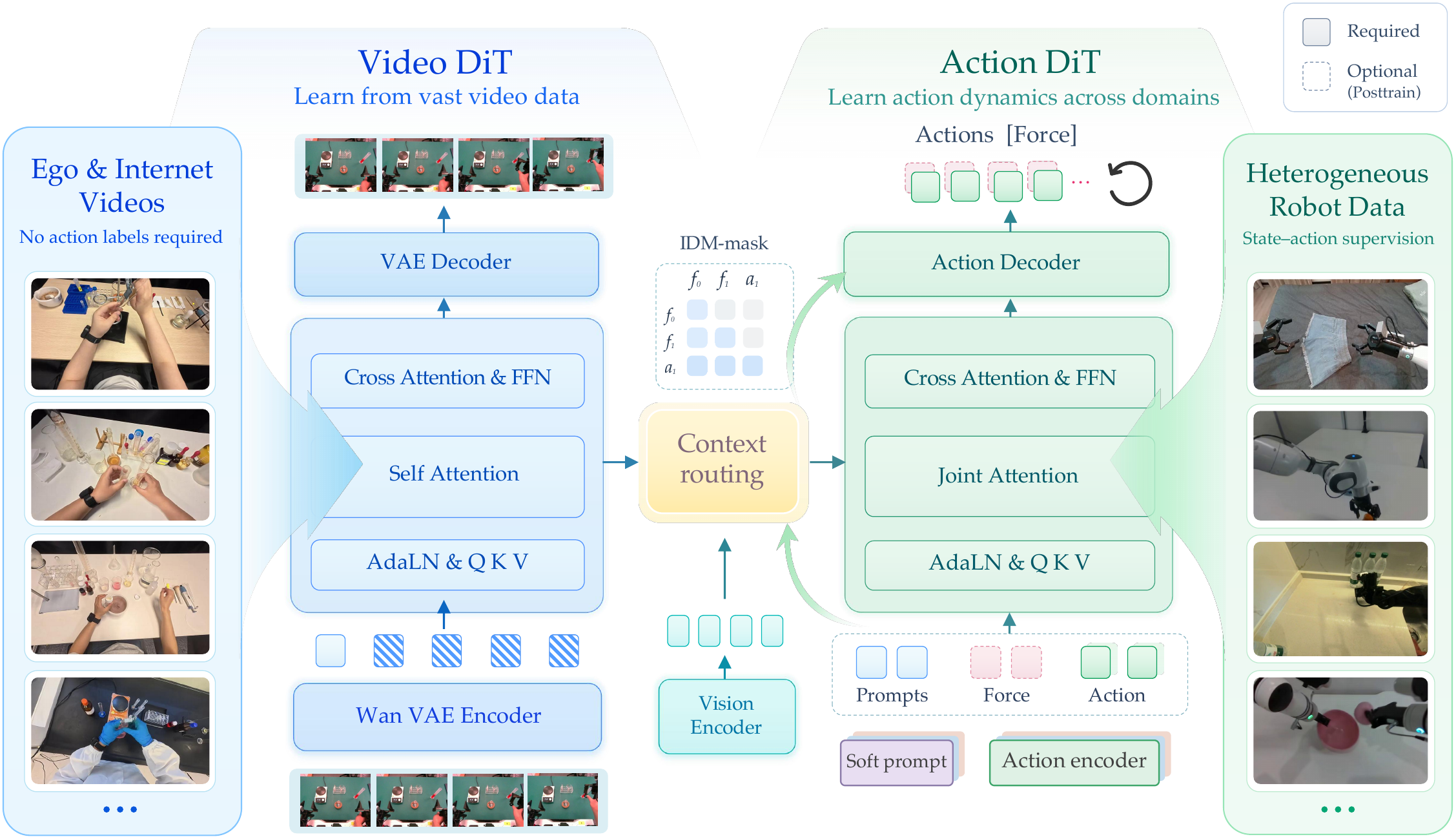}
  \caption{\textbf{Architecture of \InternWZero.}
Modality-specific video and action experts are coupled through an
observation-conditioned \emph{chunk K/V editor}. The editor adapts
cached video context to chunk-level observations, allowing
it to be reused across action updates. A frozen Wan VAE
and DINOv3 encoder provide visual representations, while precomputed
language embeddings condition both experts. Domain-specific interfaces
and soft prompts support heterogeneous embodiments; force and tactile
channels are introduced optionally during contact-aware post-training.}
  \label{fig:w0-architecture}
\end{figure}

As shown in Figure~\ref{fig:w0-architecture}, video frames are encoded
by a frozen Wan VAE~\citep{wan2025wan}, while current chunk-level
visual observations are encoded by a frozen DINOv3
encoder~\citep{simoni2026dinov}. Language instructions are represented
by precomputed text embeddings. Trainable projections connect these
representations to the corresponding experts, and proprioceptive states
provide chunk-specific conditioning for action generation. During
post-training on contact-rich tasks, force and tactile histories can be
introduced as additional action-side observations without modifying the
shared video-prediction backbone.

\paragraph{Asynchronous Duplex Inference.}
Future visual prediction provides anticipatory context, but its
computational cost makes regenerating a plan for every control update
impractical. Meanwhile, action generation must incorporate observations that arrive after a plan was produced. \InternWZero separates these
timescales: the video expert updates its prediction on a slower
schedule, while the action expert generates short chunks from the
latest available plan and current sensory feedback. After an initial
plan is generated, action inference can continue while the next video
prediction is computed.

This coordination is implemented through a cached video context
and a layerwise \emph{chunk K/V editor}. Following the
observation-guided video-context routing design of
AHA-WAM~\citep{cai2026aha}, the latest chunk-level observation is
used to adapt the cached predictive context before it is consumed by
the action expert. For each action chunk $n$, the chunk-aligned RGB
observation is first encoded by DINOv3 and projected into a visual
context. A learned query encoder summarizes this context
into a small set of routing queries.

Let $(K_\ell,V_\ell)$ denote the valid video context keys and values at
layer $\ell$. The editor first lets the observation queries $q$ retrieve
task-relevant information from the predictive video representation,
\begin{equation}
    R_{\ell,n}
    =
    \operatorname{Attn}
    \left(
        W^{q}_{\ell}q_n,\,
        K_{\ell},\,
        V_{\ell}
    \right).
\end{equation}
It then routes the retrieved information back to the original
video-token positions,
\begin{equation}
    D_{\ell,n}
    =
    \operatorname{Attn}
    \left(
        K_{\ell},\,
        R_{\ell,n},\,
        R_{\ell,n}
    \right),
\end{equation}
from which a lightweight projection predicts residual corrections
$(\Delta K_{\ell,n},\Delta V_{\ell,n})$. The chunk-specific video context is
\begin{equation}
\begin{aligned}
    \widetilde K_{\ell,n}
        &= K_\ell
        + \sigma(\gamma_\ell)\Delta K_{\ell,n},\\
    \widetilde V_{\ell,n}
        &= V_\ell
        + \sigma(\gamma_\ell)\Delta V_{\ell,n},
\end{aligned}
\end{equation}
where $\gamma_\ell$ is a learned gate. The final residual projection is
zero-initialized, so the editor begins as an identity mapping and learns
to introduce observation-dependent corrections progressively.

The underlying video context remains unchanged: each action chunk
constructs its own edited view from the same predictive context using
its newly observed visual state. Consequently, the model can preserve
longer-horizon predictive structure while adapting its local control
context as execution deviates from the previously imagined future.

\paragraph{Soft Prompts for Cross-embodiment Pretraining.}
Sharing interaction knowledge across robots requires retaining the
differences in their action semantics and sensing configurations.
Inspired by X-VLA's design of embodiment-specific prompts for
cross-embodiment learning~\citep{zheng2026x}, \InternWZero
associates each training domain $d$ with learned soft prompts
$\boldsymbol{P}^{(d)}$, prepended to every action chunk. Domains distinguish
data sources and embodiment configurations, and need not correspond
one-to-one to robot morphologies. The prompts condition shared expert
computation on these differences, while domain-specific input and
output projections adapt action and sensor representations to the
shared token space. Per-dimension validity masks distinguish missing
channels from valid zero values: unavailable inputs are zeroed before
token processing and excluded from the corresponding loss. Together,
these interfaces allow joint training without requiring identical
control or sensing semantics across robots.

\paragraph{Mixed Supervision for Prediction and Control.}

Action-free egocentric videos provide observations of physical
interactions, while action-labeled robot trajectories connect those
interactions to executable control. \InternWZero trains the same video
expert on both sources, whereas robot trajectories additionally
supervise the action expert. Real-robot and simulated trajectories
follow the same action-supervised pathway, while egocentric videos
contribute predictive supervision without requiring robot action
annotations.

Both experts use continuous flow matching. For a clean target $x_0$,
representing either video latents or a continuous action chunk, Gaussian
noise $\epsilon$, and normalized flow time $\tau\in[0,1]$, the noisy
input and target velocity are
\begin{equation}
  x_\tau=(1-\tau)x_0+\tau\epsilon,
  \qquad v^{*}=\epsilon-x_0.
\end{equation}
Flow times are sampled using the shifted continuous flow-matching
schedule, with one flow time sampled for each video example and
independent flow times for individual action chunks.
Flow times are sampled per video example and per action chunk. Video attention uses a first-frame-causal mask: the observed first
latent frame cannot attend to future video tokens, whereas future
latent frames can attend to the observed frame and interact
bidirectionally with one another. The first frame remains clean as a
temporal anchor and is excluded from the video prediction loss.
The pretraining objective combines the video and action
flow-matching losses, $\mathcal{L}_{\mathrm{video}}$ and
$\mathcal{L}_{\mathrm{joint}}$, respectively, over samples $\xi$ drawn
from the mixed training distribution $\mathcal{D}_{\mathrm{mix}}$:
\begin{equation}
  \mathcal{L}_{\mathrm{pre}}
  = \mathbb{E}_{\xi\sim\mathcal{D}_{\mathrm{mix}}}
    \left[
      \lambda_v\mathcal{L}_{\mathrm{video}}(\xi)
      +\lambda_a\,\mathbb{I}_{\mathrm{act}}(\xi)
       \mathcal{L}_{\mathrm{joint}}(\xi)
    \right],
\end{equation}
where $\lambda_v$ and $\lambda_a$ balance their contributions, and
$\mathbb{I}_{\mathrm{act}}$ indicates whether action supervision
is available. Both terms are scheduler-weighted, masked squared velocity
errors. Temporal padding, the clean current video frame, and unavailable
action dimensions are excluded from the corresponding losses.
Egocentric samples therefore contribute video supervision without
requiring action labels.

\paragraph{IDM Conditioning.}
During action-supervised training, \InternWZero maintains a second video
stream that provides predictive conditioning for the action expert.
This \emph{independent-denoising} condition stream shares all VideoDiT
parameters with the primary video flow-matching stream but uses an
independently sampled noise realization and flow time.

For each training example, the condition stream is kept clean with
probability $0.5$; otherwise, it is perturbed using an independently
sampled shifted flow time. In both cases, the first latent frame remains
clean. The primary video stream is supervised by the video
flow-matching objective, whereas the condition stream does not receive
a separate reconstruction loss. Instead, its layerwise K/V features
are consumed by the observation-conditioned editor and subsequently by
the action expert.

Importantly, the condition K/V features are not detached from the
video network. Gradients from the action flow-matching objective
therefore propagate through the K/V editor into the shared video
expert. Predictive representation learning is consequently shaped by
both future-video modeling and its usefulness for action generation,
rather than by visual reconstruction alone.

At inference time, the ground-truth condition stream is replaced by a
generated video context. Its layerwise K/V features are cached and exposed
to the action expert through the same context interface.

\paragraph{Contact-Aware Post-training with Joint Action--Contact Prediction.}
The heterogeneous pretraining stage does not require force or tactile
annotations. For downstream tasks in which contact dynamics are
critical, we extend the pretrained action interface with measured
interaction signals and jointly predict their future evolution together
with robot actions.

Visual observations alone may not fully resolve the contact conditions
that determine whether a manipulation succeeds. During contact-aware
post-training, the action expert is therefore conditioned on available
force and tactile histories, and its prediction target at physical
timestep $i$ is augmented as
\begin{equation}
  \mathbf{y}_i
  = \bigl[\mathbf{a}_i^{\top},\,
          \mathbf{f}_i^{\top},\,
          \mathbf{h}_i^{\top}\bigr]^{\top},
\end{equation}
where $\mathbf{a}_i$ is the action representation, $\mathbf{f}_i$ is the force/torque signal, and $\mathbf{h}_i$ is the tactile representation.
For each chunk, only contact history available before its generation
is supplied as conditioning; future contact channels are noised and
denoised jointly with the action channels. This formulation links
observed interaction history to both subsequent commands and their
anticipated sensory consequences. The action channels are decoded for
execution, while the contact channels represent predicted feedback.
 This extension preserves the shared predictive video backbone while
augmenting the action-side interface with task-specific physical
interaction signals.

Together, these mechanisms connect future visual context, ongoing
sensory feedback, and embodiment-specific control: predictions guide
action generation, new observations revise subsequent chunks, and
shared training links diverse interaction experience to executable
behavior.

\subsection{Training Pipeline}

The training pipeline for \InternWZero is organized as a three-layer stack that strictly separates data sampling, distributed execution, and model optimization. The \emph{data plane} is built on an in-house data engine, which curates heterogeneous robot-episode corpora into unified streams of training-ready samples. The \emph{execution plane} is built on the top of Ray~\citep{moritz2018ray}, which manages all cluster processes and co-locates data materialization with GPU-side training under explicit resource and memory budgets. The \emph{optimization plane} is implemented on top of an in-house distributed training framework: its gradient trainer owns the training loop proper, while parallelism is delegated to a pluggable backend protocol, into which \InternWZero installs the sharding, checkpointing, and compilation strategy outlined below.

\paragraph{Streaming Data Pipeline.}
To maximize data-sampling throughput of data engine, we decouple metadata from the actual sample payloads throughout the data pipeline. Instead of directly shuffling and communicating raw data, samples are first represented as compact, contiguous descriptors, and only metadata is reordered during sampling, substantially reducing communication and data-movement overhead. The corresponding payloads are materialized on demand according to the requirements of downstream training tasks. When combining heterogeneous data sources, we apply domain-level mixing weights to prevent any single corpus from dominating the training stream simply because of its raw data volume. The resulting stream is then partitioned across training nodes while preserving block order, such that batches processed locally on each node remain relatively homogeneous in their domain distribution. Sample-level random choices are generated deterministically from sample identities rather than execution or scheduling order, allowing a resumed run to reproduce exactly the same data stream. In addition, to avoid redundant I/O for video-heavy workloads, we implement a node-shared caching mechanism for video data, enabling samples already fetched and decoded by one worker to be reused across workers on the same node. This significantly reduces repeated reads and improves the overall data-loading throughput.

\paragraph{Parallelism, Precision, and Compilation.}
\InternWZero uses composable FSDP2 whose sharding units follow the model's structure: each joint mixture-of-transformers layer and each expert stage is sharded independently, while the narrow per-domain interfaces are replicated across ranks.
A single-node job uses a flat sharding mesh; in multi-node execution, the same configuration transparently promotes to a hierarchical replicate-and-shard mesh, with no changes to the model or checkpoint schema.
Mixed-precision training retains FP32 master parameters and optimizer
states, while model computation is performed in BF16. Distributed
gradient reduction is performed in FP32 to improve numerical stability. Each joint layer is compiled as a fixed-shape, full-graph unit before activation-checkpointing and sharding wrappers are applied. Selective activation checkpointing then rematerializes only a configurable prefix of layers, trading memory for recomputation without changing the objective.
Training state is persisted transactionally through distributed checkpointing, enabling exact mid-epoch resumption.

\section{Data Recipe}\label{sec:data_recipe}

The pretraining corpus of \InternWZero contains approximately 7,200 hours of data from seven datasets, spanning real and simulated robot manipulation as well as human egocentric laboratory activities. Actions and states from the robot datasets are mapped to a unified 37-dimensional representation. Numerical preprocessing, data filtering, and sampling-index construction are performed offline; during training, the pipeline reads only the resulting artifacts and precomputed normalization statistics. To balance contributions across datasets, we sample them in proportion to the square root of their numbers of effective anchors. This gives larger datasets more weight while limiting the extent to which any one dataset dominates training. Robot data cleaning targets defects that affect the supervision signals: static trajectories, intervals affected by head or waist motion, brief gripper pulses, and missing or out-of-range values. These defects are identified on each dataset's native timeline and handled with validity masks or window invalidation, retaining usable data while excluding unreliable supervision.

\paragraph{Data Composition.}
As shown in Table~\ref{tab:data-composition}, the corpus comprises four real-robot datasets, including AgibotWorld~\citep{bu2025agibot_arxiv}, Galaxea~\citep{jiang2025galaxeaopenworlddatasetg0}, RoboCOIN~\citep{wu2025robocoin}, MolmoAct~\citep{molmoact2025}, also two simulated datasets, InternData-A1~\citep{tian2026interndata} and RoboDojo~\citep{chen2026robodojo}, and one dataset of human egocentric laboratory videos (\EgoLab) collected by our team. The robot datasets include single-arm, dual-arm, and mobile manipulation. Dexterous-hand robot data and dataset copies whose format conversion is incomplete are excluded from this pretraining corpus.
We define training domains by data provenance, robot morphology, and subset identity; thus, domains do not map one-to-one to physical robot morphologies. This round comprises 25 training domains and 24 configured statistics groups, with the two Franka gripper variants sharing a single group. Of these, 23 groups provide robot-state/action normalization statistics; the remaining group corresponds to \EgoLab, which contributes only video supervision during pretraining. For language conditioning, we use a valid sub-task instruction when available and otherwise fall back to the task-level instruction.

\begin{table}
\centering
\vspace{-15pt}
\caption{\textbf{Pretraining data composition.}}
\label{tab:data-composition}
\vspace{3pt}

\small
\setlength{\tabcolsep}{4pt}

\begin{tabularx}{\linewidth}{
@{}ll
>{\raggedright\arraybackslash}Xccc@{}
}
\toprule
\textbf{Dataset} &
\textbf{Type} &
\textbf{Subsets} &
\textbf{\#Domains} &
\textbf{\#Hours} &
\textbf{\#Episodes} \\
\midrule
InternData-A1~\citep{tian2026interndata}
& Simulated
& Franka (Panda / Robotiq85), Genie1, Lift2, Split ALOHA, AC-One
& 6 & 3,494.3 & 568,194 \\
AgibotWorld~\citep{bu2025agibot_arxiv}
& Real
& gripper and tactile subsets
& 2 & 2,620.0 & 158,380 \\
RoboCOIN~\citep{wu2025robocoin}
& Real
& AgiBot G1 (2 subsets), Cobot Magic (5 subsets), Galbot G1, R1 Lite,
RMC-AIDA-L, Split ALOHA (2 subsets), Tianqin A2
& 13 & 438.1 & 59,952 \\
Galaxea~\citep{jiang2025galaxeaopenworlddatasetg0}
& Real
& R1 Lite
& 1 & 314.9 & 15,362 \\
\EgoLab
& Ego
& video-only pathway
& 1 & 275.4 & 3,192 \\
MolmoAct~\citep{molmoact2025}
& Real
& YAM bimanual
& 1 & 70.3 & 3,424 \\
RoboDojo~\citep{chen2026robodojo}
& Simulated
& dual-arm ARX X5
& 1 & 20.5 & 3,465 \\
\midrule
\textbf{Total} & & & 25 & 7,233.5 & 811,969 \\
\bottomrule
\end{tabularx}
\end{table}

\paragraph{\EgoLab Dataset.}
We collect the \EgoLab dataset to support WAMs in carrying out real-world laboratory experiments.
It comprises 275 hours of egocentric RGB video of bimanual human manipulation in real wet-laboratory settings.
These demonstrations capture manipulation priors in laboratory instrument handling and encode implicit task specifications and procedural constraints through the ordering of operations and the handling of samples and vessels.
The dataset aims at infusing this laboratory-specific operational knowledge into WAMs.

\begin{figure}[htbp]
  \centering
  \includegraphics[width=\linewidth]{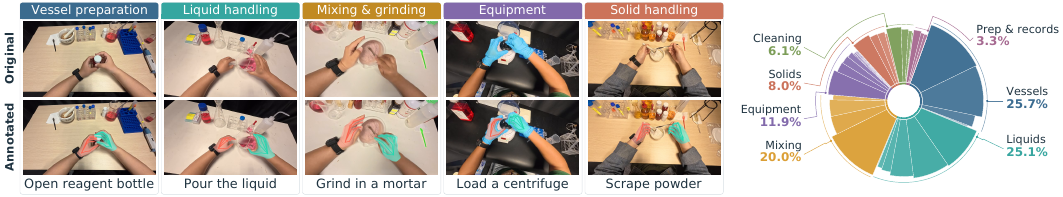}
  \vspace{-10pt}
  \caption{\textbf{\EgoLab dataset.} Left: example activities with original frames above and hand annotations below. Right: distribution of activity categories.}
  \label{fig:egolab}
\end{figure}

\textit{Video Acquisition.}
Participants perform specified laboratory tasks with selected apparatus, following task descriptions while controlling the recording with a wristband device.
A DJI Action-series camera is secured to each participant's head with a strap or helmet.
Videos are recorded at 2.5K resolution and 50\,fps and cover operations such as pipetting, sample weighing, titration, filtration, grinding with a mortar and pestle, stirring, and rinsing glassware.
Figure~\ref{fig:egolab} shows representative laboratory activities, reconstructed hand annotations, and the distribution of activity categories.

\textit{3D Motion Reconstruction.}
Camera motion and the trajectories of both hands are reconstructed at the native recording rate of 50\,fps. We use HaWoR~\citep{zhang2025hawor} for general processing and a separate reconstruction method for challenging cases. The reconstructed trajectories provide auxiliary annotations for filtering; \EgoLab is used only during \InternWZero pretraining through the video-only pathway.

\textit{Subtask Partitioning.}
We use a vision-language model accessed through a public API for open-vocabulary subtask partitioning. Frames sampled at 0.5\,fps are used to generate a global activity caption and fine-grained descriptions of visible hand--object interactions in 10\,s clips. Adjacent descriptions are merged into a subtask when they refer to the same laboratory operation, primary sample or vessel, experimental cycle, and intended outcome. The resulting partitions are mapped to training clips using the same rules as for robot datasets.

\paragraph{Unified Action and State Interface.}
Robot actions and states are mapped to a fixed 37-dimensional representation. Each arm occupies 17 dimensions: 7 for joint values, 3 for end-effector translation, 6 for orientation in the rotation-6D representation, and 1 for the gripper. The remaining 3 dimensions are reserved for the mobile base. Missing channels are zero-filled and marked by validity masks so that they do not contribute to the action loss. Head and waist degrees of freedom are outside the supervision space.

\[
\left[
\begin{array}{l}
\mathsf{left\ joints\ (7D)},\ 
\mathsf{left\ EE}\ \Delta\mathsf{pos\ (3D)},\ 
\mathsf{left\ EE}\ \Delta\mathsf{rot\ (6D)},\ 
\mathsf{left\ gripper\ (1D)}, \\[3pt]
\mathsf{right\ joints\ (7D)},\ 
\mathsf{right\ EE}\ \Delta\mathsf{pos\ (3D)},\ 
\mathsf{right\ EE}\ \Delta\mathsf{rot\ (6D)},\ 
\mathsf{right\ gripper\ (1D)},\ 
\mathsf{base}\,(\Delta x,\Delta y,\Delta\mathsf{yaw})
\end{array}
\right]
\]

Action semantics depend on the channel type. Joint actions follow each source's convention of absolute next-waypoint targets. End-effector actions encode the translation delta and relative rotation from the observation at the current control step to the next-waypoint target. Base actions encode planar displacement and yaw change over the corresponding interval. In the state representation, end-effector positions and orientations remain absolute, with orientations represented by the first two columns of the rotation matrix. Joint, gripper, and base states and actions are normalized separately. For 6D end-effectors, only the translation delta is normalized in the action representation; absolute states and action rotation-6D values remain in their raw form.

\paragraph{Temporal Sampling and Sample Construction.}
We enforce 15 fps model-time convention through temporal subsampling across all datasets. Native 60/50 fps streams use stride 4, 30/25 fps streams use stride 2, and 15 fps streams retain their original stride. For this convention, 50 fps streams are treated as 60 fps and 25 fps streams as 30 fps, while video frames are still read from their native frame positions. In stride-4 sequences, starts at frames 0, 1, 2, and 3 define four phases. Effective anchors are counted on phase-0 only; at sampling time, a legal phase for the selected anchor is chosen uniformly at random. For robot data, a legal action-window offset is then chosen within that phase. Images, states, actions, and masks remain aligned throughout each sample.
Each robot training sample contains 13 multi-view frames and 64 action steps. The multi-view images are stitched into a $384 \times 320$ canvas. Consecutive video frames are separated by 8 control steps, so the 13 frames span 96 control steps. The action sequence is divided into four 16-step chunks, each conditioned on the image and robot state at the start of that chunk. The action window is randomly offset by 0-15 steps, and no history images are used. \EgoLab follows the same video-window and temporal-sampling scheme.

\paragraph{Domain-Mixture Sampling.}
Prior robot foundation models use sublinear data reweighting to mitigate imbalances in training data: RDT-1B initializes dataset weights proportionally to the square root of dataset size~\citep{liu2024rdt}, while $\pi_0$ weights each task-robot combination proportionally to $m^{0.43}$, where $m$ is its number of training samples~\citep{black2024pi_0}.
Following this principle, we use fixed square-root weights at the domain level.
Let $N_d$ denote the number of effective phase-0 anchors in domain $d$; its target sampling probability is
\begin{equation}
p_d=\frac{\sqrt{N_d}}{\sum_j\sqrt{N_j}}.
\end{equation}
This weighting samples larger domains more frequently while reducing their dominance relative to size-proportional sampling. Each GPU batch contains samples from a single domain, although successive batches may switch domains and different GPUs may sample different domains. Domain-homogeneous batches are required by the per-domain action interfaces. Source-level target proportions are the sums of their constituent domain weights; actual consumption proportions are monitored throughout training.

\paragraph{Offline Cleaning and Filtering.}
Numerical preprocessing, filtering, and sampling-index construction are completed before training. For each source, we first standardize fields, gripper encodings, and the temporal relationship between actions and feedback to construct unnormalized 37-dimensional states and actions. We then generate legal intervals, phases, action windows, and effective-anchor indices. Training reads these derived artifacts directly. Cleaning is based on the signals used for supervision:

\begin{itemize}[leftmargin=2.2em]
  \item \textbf{Static trajectories.} We use raw motion signals to determine whether valid motion occurs and remove episodes with no valid action throughout. This test uses neither downsampled sequences nor normalized values. Trajectories containing only gripper or base motion are retained.
  
  \item \textbf{Head and waist motion.} Intervals affected by head or waist motion are excluded, along with any action window that intersects them. Other legal portions of the same episode are retained; action sequences are never concatenated across excluded intervals.
  
  \item \textbf{Gripper short pulses.} We detect \textit{open-close-open} and \textit{close-open-close} round trips on the native sequence using values normalized by the corresponding statistics group, side, and signal, before clipping. High and low signal levels are identified by normalized values of at least $0.6$ and at most $-0.6$, respectively. We then check each sampling phase to determine whether subsampling omits the intermediate reversal. If it does, the action chunk containing the event is excluded from valid supervision, and the affected window offsets are removed.
  
  \item \textbf{Missing channels and boundaries.} Missing action dimensions and temporal padding are marked with validity masks. Zero-filled missing channels do not contribute to the loss, whereas genuinely valid zeros remain supervised. Window boundaries and action validity are checked separately for each phase; validity determined for phase 0 is not assumed to hold for other phases.
\end{itemize}

\paragraph{Normalization Statistics and Split Hygiene.}
Normalization statistics are computed separately for each statistics group from values that can be sampled from the training set. The computation covers all legal phases and excludes missing channels, padding, and validation data. Gripper-pulse detection initially uses reference statistics computed from the training set. Once filtering is complete, final statistics are recomputed from the retained valid supervision.
Absolute end-effector states and rotation-6D values bypass this normalization.

\section{Simulation Benchmarks}\label{sec:simu-eval}
We evaluate \InternWZero{} in simulation to examine two complementary
properties of the pretrained model: its ability to support strong
downstream manipulation after task-specific post-training, and its
ability to retain useful physical representations when the downstream
evaluation distribution shifts beyond the demonstrations used for
adaptation.

We consider LIBERO~\citep{liu2023libero} and RoboTwin
2.0~\citep{chen2025robotwin}. LIBERO provides a standard single-arm
manipulation benchmark spanning spatial reasoning, object interaction,
goal-conditioned control, and long-horizon execution. RoboTwin 2.0
provides a broader 50-task bimanual setting and, importantly, allows us
to contrast two downstream data regimes: \emph{Full}, where both clean
and randomized demonstrations are available during post-training, and
\emph{Clean2Random}, where post-training uses only clean demonstration data and randomization is introduced only at evaluation time.

Together, these evaluations separate downstream task adaptation from
generalization beyond the downstream training distribution. We compare \InternWZero{} with representative vision-language-action (VLA) and world-action model (WAM) baselines. For each benchmark, we emphasize comparisons under the same evaluation setting and distinguish in-domain performance from out-of-domain generalization where applicable.

\subsection{Single-Arm Manipulation}
\label{sec:single_arm}

We first evaluate whether large-scale physical world-model pretraining
translates into strong task-specific manipulation performance on
LIBERO~\citep{liu2023libero}. We evaluate \InternWZero on the standard LIBERO benchmark, covering all
40 tasks across the four official task suites:
LIBERO-Spatial, LIBERO-Object, LIBERO-Goal, and LIBERO-Long.

\paragraph{Post-training setup.}
For LIBERO, we initialize from the pretrained \InternWZero{} checkpoint
and post-train the model for 50 epochs with a learning rate of
$5\times10^{-5}$. We retain the same unified 37-dimensional action
interface used during pretraining. Since LIBERO is a single-arm
manipulation benchmark, its native absolute end-effector action
representation is mapped to the corresponding left-arm slots in the
37-dimensional action space, while the unused dimensions are
zero-padded and masked out from training supervision. This allows the
shared video--action backbone to be preserved while adapting only the
valid control dimensions required by the LIBERO embodiment.

Following the benchmark protocol, we conduct 50 rollouts for each task and report
the average success rate for each suite, as well as the macro-average over
the four suites.
We compare \InternWZero against representative vision-language-action (VLA)
and world-action model (WAM) baselines, including
$\pi_0$~\citep{black2024pi_0},
$\pi_{0.5}$~\citep{pi05},
GR00T-N1.7~\citep{gr00tn1_2025},
OpenVLA-OFT~\citep{kim2025fine},
InternVLA-M1~\citep{chen2025internvla},
Xiaomi-Robotics-0~\citep{cai2026xiaomi},
Motus~\citep{bi2026motus},
LingBot-VA~\citep{li2026causal},
and Fast-WAM~\citep{yuan2026fastwam}.


\begin{table*}[t]
    \centering
    \vspace{-15pt}
    \caption{
        \textbf{LIBERO Full setting.}
    }
    \label{tab:libero_results}
    \setlength{\tabcolsep}{7pt}
    \renewcommand{\arraystretch}{1.05}
    \vspace{3pt}
    \begin{tabular}{lccccc}
        \toprule
        Method & Spatial & Object & Goal & Long & \textbf{Average} \\
        \midrule
        $\pi_0$~\citep{black2024pi_0} 
            & 98.0 & 96.8 & 94.4 & 88.4 & 94.4 \\
        InternVLA-M1~\citep{chen2025internvla} 
            & 98.0 & 99.0 & 93.8 & 92.6 & 95.9 \\
        $\pi_{0.5}$~\citep{pi05} 
            & 98.8 & 98.2 & 98.0 & 92.4 & 96.9 \\
        GR00T-N1.7~\citep{gr00tn1_2025}
            & 97.7 & 98.5 & 97.5 & 94.4 & 97.0 \\
        OpenVLA-OFT~\citep{kim2025fine} 
            & 97.6 & 98.4 & 97.9 & 94.5 & 97.1 \\
        Fast-WAM~\citep{yuan2026fastwam} 
            & 98.2 & 100.0 & 97.0 & 95.2 & 97.6 \\
        Motus~\citep{bi2026motus} 
            & 96.8 & 99.8 & 96.6 & 97.6 & 97.7 \\
        LingBot-VA~\citep{li2026causal} 
            & 98.5 & 99.6 & 97.2 & 98.5 & 98.5 \\
        \midrule
        \InternWZero 
            & 99.4
            & 99.4
            & 98.6
            & 97.0
            & 98.6 \\
        \bottomrule
    \end{tabular}
\end{table*}

\noindent \textbf{Results and Analysis.}
Table~\ref{tab:libero_results} summarizes the results on the standard
LIBERO benchmark.
\InternWZero achieves an average success rate of 98.6\%, outperforms LingBot-VA~\citep{li2026causal} and Fast-WAM~\citep{yuan2026fastwam} by 0.1 and 1.0 percentage
points, respectively.
In particular, \InternWZero achieves the best performance on
LIBERO-Spatial, reaching a success rate of 99.4\% and exceeding the
strongest reported baseline result by 0.6 percentage points.
It also obtains 99.4\% on LIBERO-Object and 98.6\% on LIBERO-Goal,
with the latter being only 0.2 percentage points below the best reported
result.
On the more challenging long-horizon suite, \InternWZero achieves a
success rate of 97.0\%.
Overall, the consistently strong performance across spatial reasoning,
object manipulation, goal-conditioned control, and long-horizon tasks
demonstrates that the pretrained model transfers effectively to
single-arm manipulation and provides a competitive foundation for
general-purpose robot control.

\subsection{Bimanual Manipulation}
\label{sec:bimanual}

We evaluate \InternWZero on bimanual manipulation using
RoboTwin 2.0.
We consider two complementary settings.
RoboTwin 2.0-Full measures in-domain multi-task performance when both
clean and randomized demonstrations are available during training,
whereas RoboTwin 2.0-Clean2Random evaluates out-of-domain generalization
from clean training environments to unseen randomized scenes.

\paragraph{Post-training setup.}
Both RoboTwin settings initialize from the pretrained
\InternWZero{} checkpoint and retain the future-IDM formulation used
during pretraining. We use a $64$-step action horizon divided into four
$16$-step chunks and a random action offset in $[0,15]$.
Chunk observations are encoded with the frozen DINOv3~\citep{simoni2026dinov} encoder, and the
action and proprioceptive signals are represented through the unified
37-dimensional interface. The native 14-dimensional ALOHA joint-action
representation is mapped to its corresponding slots in this interface,
while the remaining dimensions are zero-padded and masked out from
training supervision.
The condition-video stream is independently noised with probability
$0.5$, and the Video-DiT remains trainable so that downstream action
supervision can continue to shape the predictive representation. The model architecture and overall optimization recipe are kept aligned
between Full and Clean2Random, and both settings are post-trained for
5 epochs with a learning rate of
$2\times10^{-5}$.

\subsubsection{RoboTwin 2.0-Full}
\label{sec:robotwin_full}

We first evaluate \InternWZero{} on RoboTwin 2.0-Full, covering all
$50$ bimanual manipulation tasks.
Following the standard multi-task training protocol, the model is
fine-tuned jointly on the clean and randomized training demonstrations
and evaluated separately under the two test settings.
We report the task-averaged success rate.

\begin{table}[t]
    \centering
    \vspace{-15pt}
    \caption{\textbf{RoboTwin 2.0 Full setting.}}
    \label{tab:robotwin_full}
    \vspace{3pt}\begin{tabular}{lccc}
        \toprule
        Method & Clean (\%) & Randomized (\%) & Avg. (\%) \\
        \midrule
        $\pi_0$~\citep{black2024pi_0}             & 65.92 & 58.40 & 62.16 \\
        $\pi_{0.5}$$\pi_{0.5}$~\citep{pi05}         & 82.74 & 76.76 & 79.75 \\
        ABot-M0~\citep{yang2026abot}             & 81.20 & 80.40 & 80.80 \\
        Motus~\citep{bi2026motus}               & 88.66 & 87.02 & 87.84 \\
        Fast-WAM~\citep{yuan2026fastwam}            & 91.88 & 91.78 & 91.83 \\
        LingBot-VA~\citep{li2026causal}          & 92.90 & 91.50 & 92.20 \\
        AHA-WAM~\citep{cai2026aha}             & 93.40 & 92.20 & 92.80 \\
        OpenWAM-$\alpha$~\citep{wang2026openwam}    & 93.74 & 93.46 & 93.60 \\
        ABot-M0.5~\citep{chen2026abot}           & 94.00 & 94.20 & 94.10 \\
        \midrule
        \InternWZero        & 93.20 & 93.04 & 93.12 \\
        \bottomrule
    \end{tabular}
\end{table}

\noindent \textbf{Results and Analysis.}
Table~\ref{tab:robotwin_full} summarizes the results on
RoboTwin 2.0-Full.
\InternWZero{} achieves $93.20\%$ success in the clean setting and
$93.04\%$ under randomized evaluation, yielding an average success rate
of $93.12\%$.
The difference between the two settings is only $0.16$ percentage points,
showing that the model maintains highly stable performance when scene
randomization is introduced during both training and evaluation.

Among the WAM baselines, \InternWZero{} improves over
Fast-WAM~\citep{yuan2026fastwam} and
AHA-WAM~\citep{cai2026aha} by $1.29$ and $0.32$ percentage points
in average success, respectively, and exceeds
LingBot-VA~\citep{li2026causal} by $0.92$ points.
Its average performance is also close to OpenWAM-$\alpha$~\citep{wang2026openwam}, with a gap of
only $0.48$ points.
ABot-M0.5~\citep{chen2026abot} achieves the highest average success rate among the compared
methods at $94.10\%$, while \InternWZero{} remains within $0.98$ points.
Overall, these results show that \InternWZero{} can effectively adapt
to a diverse set of bimanual manipulation tasks while maintaining stable
performance across clean and randomized environments.

\subsubsection{RoboTwin 2.0-Clean2Random}
\label{sec:robotwin_c2r}

We further evaluate \InternWZero{} under the
RoboTwin 2.0-Clean2Random setting to measure out-of-domain generalization.
The model is fine-tuned exclusively on clean demonstrations and is then
evaluated under both clean and randomized environments.
Because randomized scene configurations are not observed during training,
this setting directly tests transfer to previously unseen visual and
environmental conditions.

\begin{table}[t]
    \centering
    \vspace{-15pt}
    \caption{\textbf{RoboTwin 2.0 Clean2Random setting.}}
    \label{tab:robotwin_c2r}
    \vspace{3pt}\begin{tabular}{lccc}
        \toprule
        Method & Clean (\%) & Randomized (\%) & Avg. (\%) \\
        \midrule
        Fast-WAM~\citep{yuan2026fastwam}
            & 77.8 & 1.9 & 39.9 \\
        X-VLA~\citep{zheng2026x}
            & 68.0 & 20.9 & 44.5 \\
        X-WAM~\citep{guo2026unified}
            & 70.0 & 25.8 & 47.9 \\
        Spatial Forcing~\citep{li2026spatial}
            & 77.2 & 26.7 & 52.0 \\
        $\pi_{0.5}$~\citep{pi05}
            & 70.7 & 46.0 & 58.4 \\
        4D-WAM~\citep{yang20264d}
            & 81.5 & 41.8 & 61.6 \\
        GigaBrain-0.7~\citep{team2026gigabrain}
            & 66.8 & 67.9 & 67.3 \\
        \midrule
        \InternWZero{}
            & \textbf{83.20} & \textbf{68.0} & \textbf{75.60} \\
        \bottomrule
    \end{tabular}
\end{table}

\noindent \textbf{Results and Analysis.}
Table~\ref{tab:robotwin_c2r} presents the results under
Clean2Random evaluation.
\InternWZero{} achieves $83.20\%$ success on clean scenes and
$68.0\%$ on randomized scenes, resulting in an average success rate of
$75.60\%$.
This is the highest average success rate among all compared methods,
exceeding GigaBrain-0.7~\citep{team2026gigabrain}, the strongest
baseline in terms of average success, by $8.30$ percentage points.
On the clean split, \InternWZero{} reaches $83.20\%$, surpassing
the strongest clean-split baseline,
4D-WAM~\citep{yang20264d}, at $81.5\%$ by $1.70$ percentage points.
Under randomized evaluation, \InternWZero{} also achieves the highest
success rate among the listed methods, marginally outperforming
GigaBrain-0.7~\citep{team2026gigabrain} at $67.9\%$ by
$0.1$ percentage points.

The resulting $15.2$-point gap between clean and randomized evaluation
indicates that unseen scene variations remain challenging.
Nevertheless, the model preserves a higher absolute
performance level than the compared baselines on both splits.
Taken together, the Clean2Random results provide evidence that
\InternWZero{} transfers effectively from clean training data to
previously unseen randomized environments.

\Needspace{30\baselineskip}
\subsubsection{Egocentric Video Pretraining with Limited Data}
\label{sec:ego_ablation}

\begin{wrapfigure}{r}{\dimexpr0.4\linewidth\relax}
    \vspace{-20pt}
    \centering
    \includegraphics[width=\linewidth]{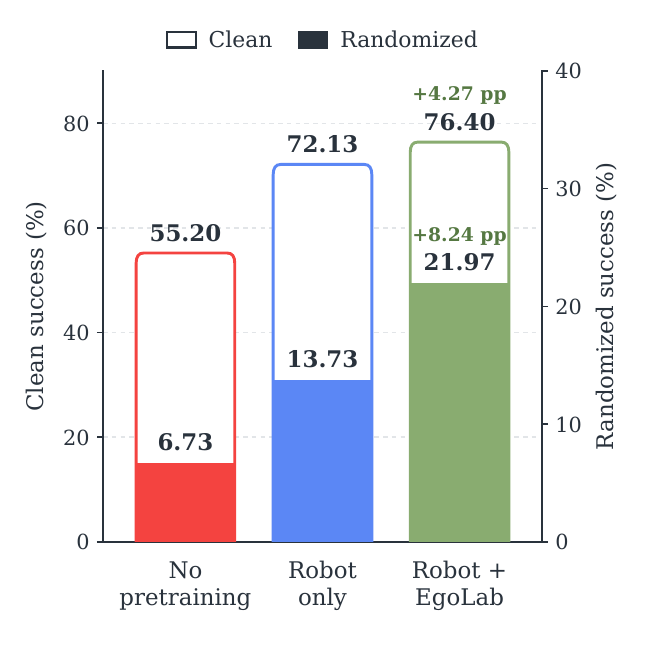}
    \caption{\textbf{Pretraining data ablation on RoboTwin C2R.}
    Outlines: Clean (left axis); solid fills: Randomized (right axis).
    Gains are relative to robot-only pretraining (pp).}
    \label{fig:ego_ablation}
\end{wrapfigure}

We investigate how egocentric video contributes to downstream
generalization under the RoboTwin 2.0-Clean2Random protocol.
This ablation uses restricted pretraining subsets, whereas the main
\InternWZero{} result in Table~\ref{tab:robotwin_c2r} uses the full pretraining corpus
We further compare three pretraining configurations: no pretraining,
robot-data-only pretraining, and joint pretraining on robot data and
\EgoLab videos.
Pretraining uses \textasciitilde{}50\,h of robot data, plus \textasciitilde{}50\,h of
\EgoLab videos for the joint variant.
All variants are subsequently fine-tuned on clean demonstrations and
evaluated on both clean and randomized environments.

\noindent \textbf{Results and Analysis.}
Figure~\ref{fig:ego_ablation} summarizes the results within this
limited-data setting.
Compared with robot-data-only pretraining, adding \EgoLab videos improves
clean success from $72.13\%$ to $76.40\%$ and randomized success from
$13.73\%$ to $21.97\%$, corresponding to gains of $4.27$ and
$8.24$ percentage points.
The larger gain on randomized scenes suggests that adding egocentric
video is particularly beneficial for generalization beyond the clean
fine-tuning distribution in this limited-data setting.
\WFclear

\section{Real-Robot and Wet-Lab Experiments}\label{sec:methodology}

\subsection{Real-Robot Experiments}
\begin{figure}[h]
    \centering
    \includegraphics[width=\linewidth]{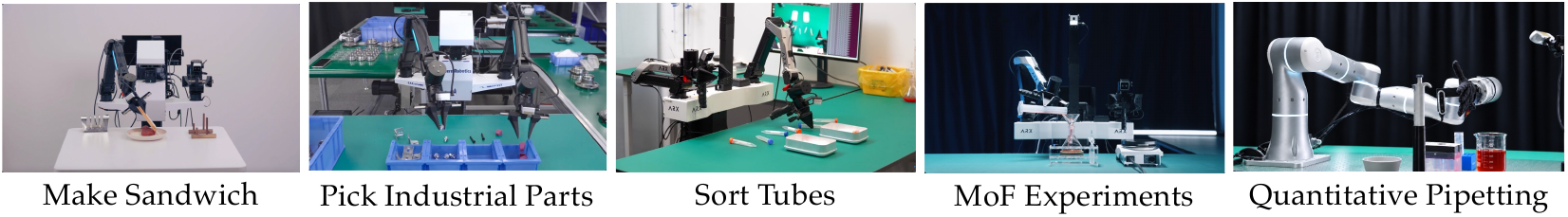}
    \caption{\textbf{Five real-world tasks.} From left to right: Make Sandwich, Pick Industrial Parts, Sort Tubes, MoF Experiments, and Quantitative Pipetting. The first four are performed on a dual-arm platform, and the last one on a dexterous-hand platform.}
    \label{fig:real_tasks}
\end{figure}

We evaluate on five real-robot tasks, shown in Figure~\ref{fig:real_tasks}, which fall into two categories: four \emph{dual-arm manipulation} tasks performed with two parallel-gripper arms, and one \emph{dexterous-hand manipulation} task performed with a multi-fingered hand mounted on a single arm.

\paragraph{Dual-Arm Manipulation.}
\textbf{Make Sandwich} is a simple long-horizon task in which the robot picks up two slices of bread and a piece of meat and stacks them in order.
\textbf{Pick Industrial Parts} requires the robot to identify each part and place it into its matching box.
\textbf{Sort Tubes} requires the robot to follow a language instruction that specifies an arm and a tube color (e.g., ``Using the right arm, pick up the orange-tipped tube and place it into the right box'') and place the matching tubes into the box on the corresponding side.
\textbf{MoF Experiments} reproduces the solution preparation stage of a metal-organic framework synthesis: the robot pours liquid from a graduated cylinder into a flask through a funnel, removes the funnel, transfers the flask onto a stirrer, inserts the stopper, and turns on the stirrer, totaling 15 sequential subtasks that require precise insertion and progress tracking across visually similar stages.

\paragraph{Dexterous-Hand Manipulation.}
\textbf{Quantitative Pipetting} evaluates dexterous manipulation and compliant tool use with a 20-degree-of-freedom (DoF) dexterous hand coupled to a 7-DoF robotic arm with hybrid force--position control. The task comprises five sequential subtasks: (i) \textit{Pickup \& reorientation}, in which the robot grasps and reorients the pipette through coordinated multi-finger in-hand manipulation; (ii) \textit{Tip attachment}, in which the robot attaches a disposable pipette tip through hybrid force--position control; (iii) \textit{Liquid aspiration}, in which the robot aspirates a prescribed volume of liquid; (iv) \textit{Liquid dispensing}, in which the liquid is dispensed into a designated test tube; and (v) \textit{Tip ejection \& return}, in which the used tip is ejected and the pipette is returned to its initial position. The task requires precise hand--arm coordination, stable contact maintenance, compliant interaction control, and reliable execution of the complete manipulation sequence.

To quantify performance, we use three complementary metrics that match the structure of each task. For every task, we run 15 trials with randomized initial object positions and orientations. For \textbf{Make Sandwich}, we report the episode-level success rate (SR), where a trial is successful only when all required actions are completed correctly. \textbf{Pick Industrial Parts} and \textbf{Sort Tubes} each present multiple target objects in a single trial, so we report the object-level success rate, i.e., the fraction of target objects that are correctly picked and placed across all trials, which measures the recall of the policy over the targets on the table. For the two multi-stage tasks, \textbf{MoF Experiments} and \textbf{Quantitative Pipetting}, all methods share the same VLM for subtask generation and we report the progress rate, computed as the proportion of subtasks completed correctly and in the prescribed order, averaged over all trials.

\begin{table}[t]
    \centering
    \caption{\textbf{Real-world results on all five tasks.} We report success rate (\%) for Make Sandwich, Pick Industrial Parts, and Sort Tubes, and progress rate (\%) for MoF Experiments and Quantitative Pipetting.}
    \label{tab:real-world_results}
    \vspace{3pt}\begin{tabular}{lccccc}
        \toprule
        \multirow{3}{*}{Method} & \multicolumn{3}{c}{Success Rate (\%) $\uparrow$} & \multicolumn{2}{c}{Progress Rate (\%) $\uparrow$} \\
        \cmidrule(lr){2-4} \cmidrule(lr){5-6}
        & \makecell{Make\\Sandwich} & \makecell{Pick Industrial\\Parts} & \makecell{Sort\\Tubes} & \makecell{MoF\\Experiments} & \makecell{Quantitative\\Pipetting} \\
        \midrule
        $\pi_{0.5}$   & \textbf{73.3} & 53.1          & 86.7 & 50.2 & 46.7 \\
        Fast-WAM       & 40.0          & 50.6          & 66.7 & 10.7 & 18.7 \\
        \InternWZero  & \textbf{73.3} & \textbf{82.7} & \textbf{88.9} & \textbf{68.4} & \textbf{65.3} \\
        \bottomrule
    \end{tabular}
\end{table}

\begin{table}[t]
    \centering
    \caption{\textbf{Subtask success rates and overall task progress rates on MoF Experiments.}}
    \label{tab:mof_subtasks}
    \vspace{3pt}\begin{tabular}{clccc}
        \toprule
        \# & Subtask & $\pi_{0.5}$ & Fast-WAM & \InternWZero \\
        \midrule
        1  & Pick up funnel from rack           & 100.0 & 100.0 & 100.0 \\
        2  & Insert funnel into flask           & 100.0 & 46.7 & 100.0 \\
        3  & Pick up graduated cylinder         & 100.0 & 13.3 & 93.3 \\
        4  & Pour liquid into flask via funnel  & 73.3 & 0.0 & 80.0 \\
        5  & Place graduated cylinder back      & 73.3 & 0.0 & 80.0 \\
        6  & Hold flask steady on platform      & 73.3 & 0.0 & 80.0 \\
        7  & Pick up funnel from flask          & 73.3 & 0.0 & 80.0 \\
        8  & Insert funnel into rack hole       & 46.7 & 0.0 & 66.7 \\
        9  & Return to initial position         & 46.7 & 0.0 & 66.7 \\
        10 & Place flask onto stirrer           & 33.3 & 0.0 & 66.7 \\
        11 & Pick up stopper from rack          & 6.7  & 0.0  & 66.7 \\
        12 & Insert stopper into flask          & 6.7  & 0.0  & 66.7 \\
        13 & Press left button of stirrer       & 6.7  & 0.0  & 26.7 \\
        14 & Press right button of stirrer      & 6.7  & 0.0  & 26.7 \\
        15 & Return to initial position         & 6.7  & 0.0  & 26.7 \\
        \midrule
        \multicolumn{2}{l}{\textbf{Progress rate}} & 50.2 & 10.7 & \textbf{68.4} \\
        \bottomrule
    \end{tabular}
    \vspace{-10pt} 
\end{table}

\begin{table}[t]
    \centering
    \small
    \caption{\textbf{Subtask success rates and overall task progress rates on Quantitative Pipetting task.}}
    \label{tab:pipetting_results}
    \vspace{3pt}\begin{tabular}{clccc}
        \toprule
        \# & Subtask & $\pi_{0.5}$ & Fast-WAM & \InternWZero \\
        \midrule
        1  & Pickup \& reorientation             & \textbf{93.3} & 46.7 & 86.7 \\
        2  & Tip attachment             & 40.0 & 13.3 & \textbf{66.7} \\
        3  & Liquid aspiration          & 33.3 & 13.3 & \textbf{66.7} \\
        4  & Liquid dispensing    & 33.3 & 13.3 & \textbf{60.0} \\
        5  & Tip ejection \& return       & 33.3 & 6.7 & \textbf{46.7} \\
        \midrule
        \multicolumn{2}{l}{\textbf{Progress rate}} & 46.7 & 18.7 & \textbf{65.3} \\
        \bottomrule
    \end{tabular}
\end{table}

As shown in Table~\ref{tab:real-world_results}, \InternWZero matches or outperforms $\pi_{0.5}$ on all five tasks and outperforms Fast-WAM by a large margin.
On \textbf{Make Sandwich}, both pretrained policies reach 73.3\%, as basic pick-and-stack behaviors are already well covered by pretraining.
On \textbf{Pick Industrial Parts}, \InternWZero improves the object-level success rate from 53.1\% to 82.7\%: most failures of $\pi_{0.5}$ are recognition errors where a part is placed into the wrong box, which we attribute to the stronger part-level discrimination provided by the visual pretraining of the WAM backbone.
On \textbf{Sort Tubes}, both pretrained models follow the instruction reliably (88.9\% vs.\ 86.7\%), while Fast-WAM (66.7\%) often closes the gripper above the thin tubes, as its visual representation trained from scratch fails to localize the correct grasp point.

The benefit of prediction is clearest on \textbf{MoF Experiments}, where \InternWZero improves the progress rate from 50.2\% to 68.4\%.
The subtask breakdown in Table~\ref{tab:mof_subtasks} reveals two failure modes of $\pi_{0.5}$.
First, at the pouring step (subtask~4), $\pi_{0.5}$ misaligns the graduated cylinder with the funnel and spills liquid, whereas \InternWZero pours more steadily. Second, $\pi_{0.5}$ degrades sharply from subtask~8 onward (46.7\% $\rightarrow$ 33.3\% $\rightarrow$ 6.7\%), where the small transparent funnel, flask, and stopper must be precisely grasped and inserted, while \InternWZero sustains 66.7\% through subtask~12 and only drops at the final button-pressing stage (26.7\%), which demands millimeter-level precision.
Fast-WAM performs poorly on this task: without pretraining, it suffers from severe state confusion across the visually similar stages, repeatedly re-executing or skipping steps, and never progresses beyond the first three subtasks.
Its results, which trail \InternWZero by a large margin on every task, confirm that the gains of \InternWZero stem from pretraining and the predictive objective rather than from the architecture alone.

To evaluate whether \textbf{force-aware active hybrid force--position interaction} enables reliable dexterous manipulation and compliant tool use, we use the quantitative pipetting task as a challenging testbed, with the results reported in Table~\ref{tab:pipetting_results}. It requires a 20-DoF dexterous hand to achieve stable grasping and precise in-hand reorientation of the pipette through coordinated multi-finger movements. Moreover, tip attachment and tip ejection and return require human-like active hybrid force--position interaction, in which the robot regulates contact forces while maintaining the desired tool pose during contact.  Among the evaluated methods, \InternWZero achieves the strongest overall performance, obtaining the highest success rates on four of the five subtasks and the highest overall progress rate of 65.3\%. Its advantages are particularly pronounced on the contact- and force-sensitive subtasks, including tip attachment, liquid aspiration, liquid dispensing, and tip ejection. Although \(\pi_{0.5}\) performs better on pickup reorientation, \InternWZero substantially outperforms the other methods in the subsequent manipulation stages. These results suggest that reliable quantitative pipetting benefits from the combination of high-DoF dexterous manipulation and active hybrid force–position interaction.

\begin{figure}[h]
    \centering
    \includegraphics[width=\linewidth]{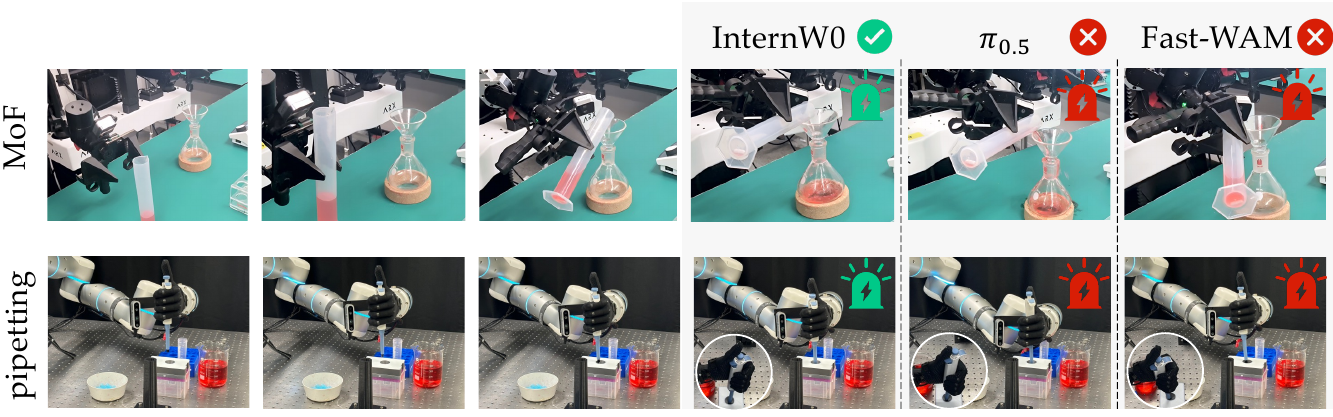}
    \caption{Qualitative comparison on the pouring step of MoF Experiments (top) and the tip-attachment step of Quantitative Pipetting (bottom). \InternWZero aligns the tool before executing the contact-rich action, whereas the baselines act with visible misalignment.}
    \label{fig:failure_cases}
\end{figure}

\paragraph{Failure Case Analysis.}
Figure~\ref{fig:failure_cases} visualizes the two most representative failure modes.
In the pouring step of \textbf{MoF Experiments}, \InternWZero keeps the mouth of the graduated cylinder centered over the funnel while tilting, so the liquid enters the flask cleanly; $\pi_{0.5}$ starts tilting slightly off-center and part of the liquid runs down the outside of the flask, while Fast-WAM tilts the cylinder far from the funnel and pours directly onto the platform. In the tip-attachment step of \textbf{Quantitative Pipetting}, through force-aware active hybrid position--force interaction, \InternWZero accommodates an initially misaligned pipette by adapting its pose online in response to contact forces and yielding compliantly to external forces, thereby aligning the tip with the target opening and successfully lifting the pipette. In contrast, $\pi_{0.5}$ continues to apply downward force during insertion until the pipette slips out of its grasp, whereas Fast-WAM exhibits a substantial positional error at the insertion site, bending and deforming the pipette body.
\section{Efficiency}
\label{sec:efficiency}

For a physical world model, predictive accuracy alone is insufficient:
predictions must also become available in time to influence ongoing
interaction. This makes \emph{timeliness} a model-level requirement,
rather than merely a systems consideration.

\InternWZero addresses this requirement by separating the
computational timescales of prediction and control. A high-capacity
Video-DiT maintains longer-horizon predictive context at a slower
timescale, while a lightweight Action-DiT updates actions more
frequently from the latest observation and available predictive
context. Together with the observation-conditioned context-routing
interface introduced in Section~\ref{sec:w0-architecture}, this design
keeps expensive future prediction off the latency-critical action path
without removing its influence on control.

\subsection{Critical-Path Latency and Policy Update Rate}
\label{sec:latency}

\begin{wrapfigure}{r}{0.5\textwidth}
\centering
\vspace{-10pt}
\includegraphics[width=0.49\textwidth]{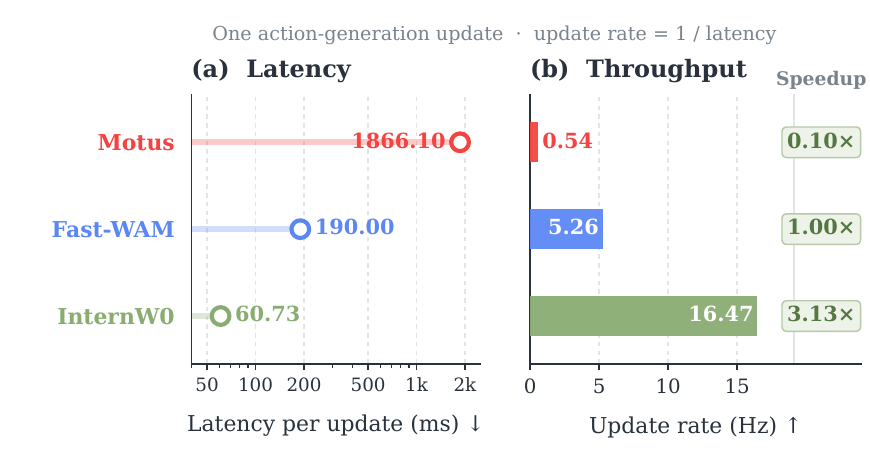}
\vspace{-10pt}
\caption{\textbf{Critical-path inference efficiency.}}
\label{fig:w-efficiency}
\end{wrapfigure}

We evaluate inference efficiency using the same action-generation
protocol across all compared methods. For \InternWZero, we distinguish
the \emph{critical action path} from the slower predictive path. The
critical-path latency includes current-observation encoding,
observation-conditioned context routing, and Action-DiT denoising using
the latest available predictive context. Video-plan generation is
scheduled asynchronously and is therefore not included in this
critical-path measurement.

Conceptually, under asynchronous execution, the rate of closed-loop
policy updates is determined by the latency of the fast action path,
whereas Video-DiT inference belongs to the slower predictive path and
does not directly block each action update. All latency measurements
are performed on the same RTX 5090D hardware under the same numerical
precision setting.


\noindent\textbf{Results and Analysis.}
As shown in Figure~\ref{fig:w-efficiency}, \InternWZero reduces the
critical-path action-generation latency to $60.73$ ms, corresponding
to a maximum model-side policy update rate of $16.47$ Hz. This is a
$3.13\times$ speedup over Fast-WAM and substantially reduces latency
relative to Motus.

The practical significance of this reduction is not only higher
throughput. A shorter critical path decreases the interval between
receiving a new physical observation and producing an action update
conditioned on that observation. In closed-loop interaction, this directly affects how quickly the policy can respond when the actual environment begins to deviate from its predicted evolution.

\subsection{Asynchronous Prediction and Closed-Loop Execution}
\label{sec:async_inference}

Following the asynchronous planner--executor principle introduced in
AHA-WAM~\citep{cai2026aha}, \InternWZero decouples low-frequency
world prediction from high-frequency action generation. In
\InternWZero, this principle is instantiated through the pretrained
Video-DiT and Action-DiT together with the observation-conditioned
video-context interface described in
Section~\ref{sec:w0-architecture}.

The Video-DiT periodically produces a longer-horizon predictive plan
and exposes its layerwise K/V representations as reusable predictive
context. Between video-plan updates, the Action-DiT does not rerun the
full video predictor. Instead, each action update reads the latest
available video context and adapts it using the newly observed visual
state through the chunk K/V editor before generating the next action
chunk. When a refreshed video plan becomes available, it replaces the
previous predictive context for subsequent action updates.

This organization moves the expensive video-generation process away
from the latency-critical control path. The Video-DiT determines how
often long-horizon predictive context can be refreshed, whereas the
Action-DiT determines how quickly new observations can influence
control. The two frequencies serve different purposes: prediction maintains longer-horizon physical context, while the fast action path maintains responsiveness to the realized state of the environment.

Importantly, asynchronous execution does not imply that the predictive plan is treated as fixed between updates. Observation-conditioned K/V routing continuously changes how the cached plan is exposed to each action chunk. The system can reuse expensive predictive computation while remaining sensitive to discrepancies between the predicted future and the state that unfolds.

\paragraph{Critical-path acceleration.}
Beyond asynchronous scheduling, the deployment pipeline reduces the
cost of the fast action path through complementary implementation
optimizations. Graph-level acceleration is applied to the Action-DiT
and context-processing modules to reduce launch and execution overhead,
while selective compilation is used for the Video-DiT prefill path.
Computations invariant across denoising steps are hoisted outside the
inner sampling loop, and redundant tensor transfers and repeated state
traversal are removed.

These optimizations do not alter the model architecture or inference
precision. Instead, they reduce systems overhead around the same
video-context routing and action-generation computation, further
separating the latency of responsive control from the cost of
long-horizon prediction.

\paragraph{Efficiency as a property of physical intelligence.}
Taken together, asynchronous scheduling, reusable predictive context,
observation-conditioned routing, and critical-path acceleration allow
\InternWZero to retain an expressive video predictor without making
its full generation latency the bottleneck of every action update.
The resulting interaction loop continuously observes, updates, and
acts while predictive computation proceeds at its own timescale.

This closes the loop with the design motivation of the InternW series:
a useful physical world model should not only predict accurately and
produce decision-relevant representations, but should make that
information available before it becomes stale. This requirement is
particularly important in dynamic and contact-rich interactions, where
the value of a prediction depends not only on what it predicts, but
also on whether the agent can react to the physical evidence that
arrives afterward.
\section{Summary and Future Work
}\label{sec:future-directions}
In this report, we presented \InternWZero,
which establishes a foundation for physical world modeling by combining visual and force perception with language, proprioception, prediction, and action. Its duplex interaction, enabled by asynchronous, multi-frequency processing supports continuous sensing, prediction, action, and correction, and makes \InternWZero work much efficiently. On mainstream simulation benchmarks of robot manipulation, \InternWZero has shown its highly competitive performance. Furthermore, we apply \InternWZero to real-world scientific experiments that require dexterous, precise, and safe manipulation. 

Future work will extend this foundation in the following directions.

\paragraph{From Observation to Intervention Understanding.}
\InternWZero integrates force feedback with vision and body state, helping robots determine whether objects are grasped securely, contact is established, or an operation has reached the desired state. Future models should incorporate additional physical modalities, including tactile sensing, torque, proprioception, and contact signals. They should learn not only what happens, but how interventions change subsequent states, distinguishing passive evolution from action-induced effects and updating hidden-state estimates as new feedback arrives. This capability is essential for scientific procedures whose outcomes depend on timing, force, temperature, mixing, and operation order \citep{scholkopf2021toward}.

\paragraph{Broader Modalities and Specialized Experts.}
Future \InternW models should incorporate additional modalities, including depth, 3D geometry, audio, and other task-specific observations, through modality-specific encoders and decoders. These experts should exchange information through a shared physical representation while preserving modality-specific temporal and spatial structure. Realizing this extension requires multimodal data with reliable timestamps and spatial calibration. The data pipeline should filter sensor failures, corrupted intervals, implausible trajectories, and annotation errors, normalize units and coordinate conventions, and record validity masks for missing channels. 

\paragraph{Prediction Across Scales and Horizons.}
Physical processes combine fast local dynamics with slower, multi-stage changes. Future models should connect these scales through hierarchical representations, memory, and adaptive temporal resolution. \InternWZero's asynchronous, multi-frequency design provides a basis for long-horizon prediction and replanning while maintaining high-frequency control. Reliability will require handling partial observability, accumulated error, and intervention-dependent dynamics through uncertainty-calibrated prediction, continual state correction, and evaluation across multiple timescales \citep{kim2019variational}.

\paragraph{One Physical Foundation for Many Environments.}
The \InternW world model series should transfer across robots, instruments, laboratories, and physical settings while allowing rapid specialization. Building on \InternWZero's heterogeneous data pipeline and domain-specific action interfaces, future models should combine shared physical representations with adaptable interfaces for sensors, actuators, embodiments, and action spaces. Parameter-efficient adaptation and limited demonstrations should enable specialization to new robots and protocols without relearning basic physical knowledge. Evaluation should measure transfer performance and the data and interaction required for adaptation.

\paragraph{Closed-Loop Operation in the Real World.}
A physical world model must predict, act, observe the outcome, and update its next decision. \InternWZero's duplex and asynchronous architecture provides a foundation for background planning alongside responsive action generation. Real-world prediction errors can reveal missing data, calibration problems, and environment-specific failure modes. Future systems should use these signals for continual improvement while preserving existing capabilities \citep{kirkpatrick2017overcoming}. Evaluation should include sustained task success, disturbance recovery, transfer, latency, and reliable action correction, especially in multi-stage wet-lab procedures.

\paragraph{Lightweight World Models for Embedded Intelligence.}
Future \InternW world models should target approximately 4B parameters and ultimately fewer than 1B, enabling deployment on customized small devices and embedded physical agents so as to work in various physical environments. Distillation, quantization, compact latent representations, efficient action heads, and hardware-aware inference will be important. Compression should preserve multimodal state estimation, intervention-sensitive prediction, high-frequency action generation, and recovery from unexpected contact, while accounting for memory, energy, and perception costs. Combined with rapid specialization, lightweight \InternW models could serve as local world models for diverse robots and scientific instruments.

\section*{Contributors}

The following are the contributors of this article:
Jisong Cai, Yao Mu, Ganlin Yang, Zhe Cao, Zhangzheng Tu, Xing Gao, Kailin Li, Xinyu Zhan, Lixin Yang, Yangkun Zhu, Haoxiang Ma, Ming Zhou, Qiaojun Yu, Yufei Xue, Liqun He, Yifei Yao, Yifan Zhu, Long Ling, Bingqi Jiang, Haoyu Guo, Xueyue Zhu, Bowen Zhou, Bin Zhao, Tianfan Xue, Chunhua Shen, and Weinan Zhang.

\bibliographystyle{plainnat}
\bibliography{references}
\end{document}